\documentclass[letterpaper, journal, twoside]{IEEEtran}

\usepackage{cite}
\usepackage{amsmath,amssymb,amsfonts}
\usepackage{graphicx}
\usepackage{textcomp}
\usepackage{xcolor}
\usepackage{booktabs}
\usepackage{multirow}
\usepackage{array}
\usepackage[hyphens]{url}
\usepackage{microtype}
\usepackage{hyperref}
\usepackage{needspace}
\usepackage{flafter}
\usepackage{caption}     
\usepackage{placeins}    

\begin{document}

\markboth{Preprint Version. April, 2026}
{Irfan \MakeLowercase{\textit{et~al.}}: Knee-xRAI: An Explainable AI Framework for Kellgren--Lawrence Grading of Knee Osteoarthritis}

\title{Knee-xRAI: An Explainable AI Framework for Automatic Kellgren--Lawrence Grading of Knee Osteoarthritis}

\author{Azmul~A.~Irfan$^{1\star}$, 
        Nur~Ahmad~Khatim$^{2}$, 
        Alfan~Alfian~Irfan$^{3}$, 
        Achmad~Zaki$^{1}$, 
        Erike~A.~Suwarsono$^{1}$,
        and~Mansur~M.~Arief$^{4}$%
\thanks{$^1$Azmul~A.~Irfan and Achmad~Zaki, are with Orthopaedic Department, Faculty of Medicine UIN Syarif Hidayatullah Jakarta, Indonesia (e-mail: azmul.irfan@uinjkt.ac.id).}%
\thanks{$^2$Nur~Ahmad~Khatim is with Informatics Engineering, Institut Teknologi Sepuluh Nopember, Indonesia}%
\thanks{$^3$Alfan~Alfian~Irfan is with Information Technology, Universitas Muhammadiyah Yogyakarta, Indonesia.}%
\thanks{$^4$Mansur~M.~Arief is with Industrial and Systems Engineering, King Fahd University of Petroleum and Minerals, Saudi Arabia.}%
\thanks{$^5$, Erike~A.~Suwarsono is with Microbiology Department, Faculty of Medicine UIN Syarif Hidayatullah Jakarta, Indonesia}
\thanks{$^{\star}$Corresponding Author}%
\thanks{Code available at: \url{https://github.com/ois-lab/knee-xrai}}%
}%

\maketitle

\begin{abstract}
Grading knee osteoarthritis (KOA) on plain radiographs is poorly reproducible across readers. Because a single-grade disagreement on the Kellgren--Lawrence (KL) scale can alter surgical management or redirect a patient from conservative therapy to intra-articular injection, clinical decision-making is highly sensitive to reader variability. Meanwhile, deep learning models that outperform human readers often offer no explanation for their decisions. We present Knee-xRAI, a pipeline that decomposes the grading process by mimicking clinical radiological workflows. It independently measures joint space narrowing (JSN), osteophytes, and subchondral sclerosis, and then combines these findings into an explainable KL grade. Specifically, a U-Net++ architecture quantifies JSN via contour segmentation, an SE-ResNet-50 multi-task network grades osteophytes per anatomical site on the OARSI scale, and a hybrid texture-CNN detects binary sclerosis. This pipeline yields a 50-dimensional feature vector evaluated via an XGBoost-SHAP classifier (Path~A, audit) and a ConvNeXt hybrid predictor (Path~B, deployed). On 8,260 OAI-derived radiographs, the JSN module achieved a Dice score of 0.8909 and an mJSW ICC of 0.8674. Path~A reached a QWK of 0.6294 and an AUC of 0.8046, confirming that the structured feature vector alone carries a substantial diagnostic signal. Path~B achieved a QWK of 0.8436 and an AUC of 0.9017. SHAP analysis identifies JSN as the dominant feature, with osteophytes adding a consistent increment and sclerosis contributing marginally. Ablation experiments demonstrate that the predictor genuinely relies on these measurements. Removing JSN evidence collapses KL3--KL4 recall while early grades remain intact, aligning precisely with the diagnostic criteria of the KL scale. Knee-xRAI grounds every prediction in an auditable chain of measured radiographic findings, providing clinical transparency at the point of care.
\end{abstract}

\needspace{6\baselineskip}

\section{Introduction}
\IEEEPARstart{K}{nee} osteoarthritis (KOA) affects approximately 654 million adults aged 40 years and older~\cite{cui2020global} and remains a leading cause of lower-limb disability worldwide~\cite{steinmetz2023global}. This burden is disproportionately high in low- and middle-income health systems. In Indonesia, for example, a population of 26.8 million adults aged 60 years or older~\cite{bps2019} is served by only 1.2 radiologists per 100,000 people~\cite{yunus2024radiology}. Consequently, radiographic interpretation is routinely delegated to non-subspecialist clinicians in these settings. Interpretable diagnostic support, whose underlying logic can be audited without expert radiological intervention, is therefore a critical clinical requirement rather than a theoretical luxury, aligning with recent deployment-focused AI efforts in healthcare settings~\cite{reyblanes2024knee}.

Radiographic severity follows the Kellgren--Lawrence (KL) scale, a five-grade system (0--4) based on joint space narrowing (JSN), osteophytes, and subchondral sclerosis~\cite{kellgren1957radiological}. Significant inter-reader variability, driven by divergent KL descriptions in the literature~\cite{schiphof2008differences}, carries severe clinical consequences. A single-grade disagreement can prematurely redirect a patient from conservative therapy to intra-articular corticosteroid injections, or inappropriately alter eligibility for orthopaedic referral~\cite{culvenor2015defining}.

While deep learning models achieve high performance on the KL scale, reaching quadratic weighted kappa (QWK) values up to 0.83~\cite{tiulpin2020automatic}, they typically compress features into opaque, end-to-end predictions without an auditable link to specific clinical findings~\cite{weber2024xai}. Partial solutions exist, for example, MediAI-OA~\cite{yoon2023assessment} quantifies JSN and osteophytes but omits sclerosis, diverging from standard radiological workflows. Currently, no published framework concurrently quantifies all three KL-defining features within a single, auditable grading pipeline.

This study addresses this gap by introducing Knee-xRAI, a deep learning pipeline that explicitly quantifies all three KL structural components from plain radiographs to support ordinal classification. We hypothesize that this explicit design can match end-to-end classification performance while substantially improving clinical interpretability by providing an auditable feature surface. Our primary contributions include: (i)~the Knee-xRAI pipeline, which concurrently quantifies all three foundational KL features (JSN, osteophytes, and subchondral sclerosis) within a single workflow (Table~\ref{tab:relatedwork}); (ii)~a two path architecture comprising a deployed ConvNeXt hybrid predictor (Path~B, QWK 0.8436) and an XGBoost/SHAP audit instrument (Path~A); (iii)~inference-time ablation confirming that the predictor's internal reliance hierarchy mirrors KL diagnostic logic; and (iv)~the public release of our source code, pretrained weights, and per-site annotations at \url{https://github.com/ois-lab/knee-xrai}.

\section{Related Work}

AI-assisted KOA grading comprises end-to-end classification, feature-specific quantification, and explainability methods. We briefly review these domains to highlight the gap Knee-xRAI addresses.

\subsubsection{End-to-End KL Classification.}
Most paradigms predict global KL grades directly from radiographs. Early frameworks used YOLOv2\,+\,CNN combinations ($69.7\%$ accuracy)~\cite{chen2019fully}, which evolved into CNN ensembles~\cite{pi2023ensemble,pan2024automatic} and the state-of-the-art VL-OrdinalFormer~\cite{ullah2026vl} (ViT-L/16, CORAL ordinal regression, and CLIP visual--language alignment). However, these black-box architectures do not isolate the specific structural findings driving predictions, limiting clinical adoption where verification is mandatory.

\subsubsection{Feature-Specific Quantification.}
JSN is the primary marker of advanced disease and cartilage integrity~\cite{kellgren1957radiological}, while subchondral sclerosis serves as a late-stage confirmatory sign~\cite{kellgren1957radiological}. For JSN, automated tracking is established via TransUNet ($\text{Dice} = 0.889$, $\text{ICC} = 0.927$)~\cite{guo2025predicting} and continuous minimum joint space width (mJSW) normalization~\cite{yoon2023assessment}. For osteophytes, Tiulpin and Saarakkala~\cite{tiulpin2020automatic} achieved per-site OARSI kappas of 0.79--0.94 using an SE-ResNet-50. Sclerosis detection foundations rely on deep networks (EfficientNet-B0, $\text{AUC} = 0.952$)~\cite{kim2025classification} and handcrafted regional texture descriptors ($\text{AUC} = 0.840$)~\cite{bayramoglu2020adaptive}.

\subsubsection{Explainability and Research Gap.}
Grad-CAM~\cite{selvaraju2020grad} is widely used but often suffers from an \textit{interpretation gap} where heatmaps fail to faithfully represent true model reliance~\cite{weber2024xai}. While concept-based models like VL-OrdinalFormer~\cite{ullah2026vl} embed clinical features implicitly, Knee-xRAI quantifies each feature explicitly. As shown in Table~\ref{tab:relatedwork}, no prior framework concurrently quantifies all three foundational KL features. Knee-xRAI bridges this gap via the four-stage pipeline detailed below.

\begin{table*}[!t]
\caption{Comparison of existing KOA AI systems with Knee-xRAI.}
\label{tab:relatedwork}
\centering
\begin{tabular*}{\textwidth}{@{\extracolsep{\fill}}lccccl}
\toprule
System & JSN & Osteo. & Sclerosis & KL & XAI Type \\
\midrule
Chen et al.~\cite{chen2019fully}        & --       & --       & --       & \checkmark & None \\
Pi et al.~\cite{pi2023ensemble}         & --       & --       & --       & \checkmark & Grad-CAM \\
MediAI-OA~\cite{yoon2023assessment}     & Quant.   & Quant.   & --       & \checkmark & Feature-level \\
Kim et al.~\cite{kim2025classification} & --       & --       & Quant.   & --         & Grad-CAM \\
VL-OrdinalFormer~\cite{ullah2026vl}     & Implicit & Implicit & Implicit & \checkmark & CLIP maps \\
\textbf{Knee-xRAI}                      & Quant.   & Quant.   & Quant.   & \checkmark & Multi-layer \\
\bottomrule
\end{tabular*}
\end{table*}

\section{Methods}
This section describes the dataset and evaluation protocol, then details the four-stage Knee-xRAI pipeline from JSN segmentation through KL grade classification. 

\subsection{Dataset, Preprocessing, and Evaluation Protocol}

Experiments utilized the public Knee Osteoarthritis Severity Grading Dataset~\cite{chen2018knee,chen2019fully}, a processed subset of the Osteoarthritis Initiative (OAI) repository. It contains 8,260 grayscale posteroanterior knee radiographs ($224 \times 224$ pixels) partitioned via a 70:10:20 stratified split into training ($n = 5,778$), validation ($n = 826$), and test ($n = 1,656$) sets. The class distribution is heavily skewed toward lower severity: KL0 is the majority class ($39.4\%$, $n = 3,253$), while KL4 is rare ($3.0\%$, $n = 251$).

Feature-level ground truth was compiled across subsets of 400--500 images, independently reviewed by two orthopaedic specialists (inter-annotator agreement was not quantified due to resource constraints, see Section~\ref{sec:limitations}). Annotations included: (i)~polyline articular contours for JSN; (ii)~OARSI ordinal grades (0--3) across four anatomical sites following the OARSI atlas~\cite{altman1995atlas}; and (iii)~binary subchondral sclerosis labels (absent vs. present). The test partition was strictly withheld until final evaluation. Using a de-identified public repository~\cite{chen2018knee}, institutional review board (IRB) approval was waived.

Images were preprocessed using CLAHE (clip limit 3.0, $8 \times 8$ pixel tile grid), intensity-clipped at the 5th/99th percentiles, normalized to $[0,1]$, and standardized via ImageNet statistics ($\mu = 0.485$, $\sigma = 0.229$). Training augmentations included random rotations ($\pm 10^\circ$), horizontal flips ($p = 0.5$), brightness/contrast jitter ($\pm 15\%$), and affine scaling ($[0.9, 1.1]$). Class imbalance was addressed via inverse-frequency batch sampling and an ordinal soft-label cross-entropy loss with Gaussian smoothing ($\sigma = 0.5$). For osteophytes, full-image CLAHE preceded $140 \times 140$ pixel patch extraction but was omitted during subsequent ROI processing to avoid artifact accumulation. For sclerosis, CLAHE was entirely omitted to preserve the natural subchondral intensity distribution required by the handcrafted texture descriptors (GLCM and fractal dimensions).

The JSN module was evaluated using Dice, HD95, and ICC against manual mJSW measurements. Osteophyte grading used per-site QWK ($\kappa$) and AUC, while sclerosis classification used macro $F_1$, balanced accuracy, and AUC. Global KL prediction was evaluated primarily via QWK, with accuracy, macro $F_1$, and macro AUC as secondary metrics. Two ablation studies quantified feature-family contributions (Path~A) and the structured pathway's marginal utility (Path~B). Statistical $95\%$ confidence intervals were computed via percentile bootstrap (1,000 resamples). Models were implemented in PyTorch 2.0 via PyTorch Lightning on a single NVIDIA RTX 4090 GPU.

\subsection{Knee-xRAI Framework}

\begin{figure*}[!t]
\centering
\includegraphics[width=\textwidth]{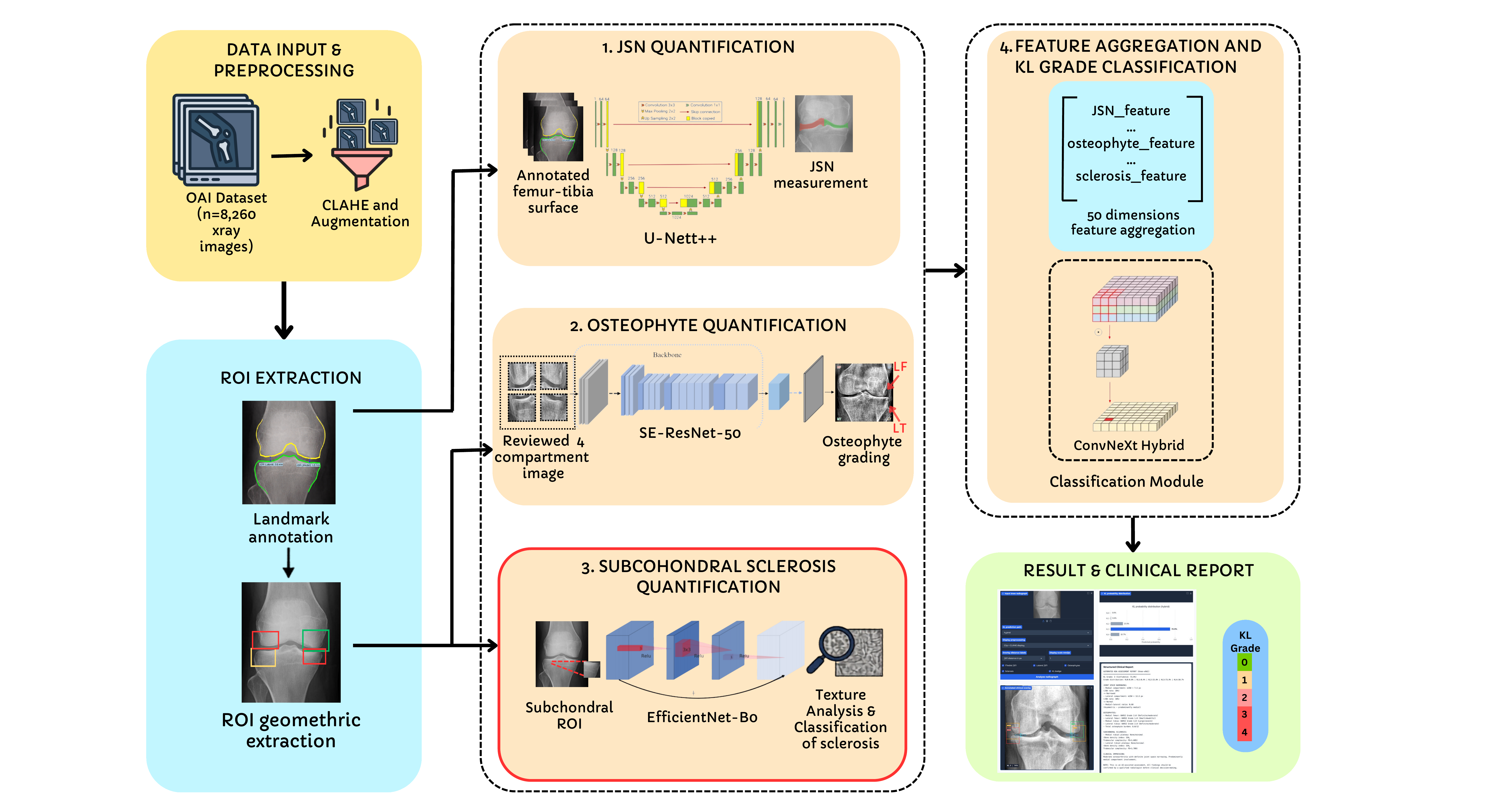}
\caption{Overview of the Knee-xRAI four-stage pipeline.}
\label{fig:pipeline}
\end{figure*}

Knee-xRAI decomposes KL grading into four sequential stages
(Fig.~\ref{fig:pipeline}).

\subsubsection{Stage 1: JSN Segmentation and Quantification.}
A U-Net++~\cite{zhou2018unetplusplus} with an ImageNet-pretrained EfficientNet-B4 encoder~\cite{tan2020efficientnet} was trained to generate three-class masks (background, medial, and lateral joint spaces) from $224 \times 224$ px radiographs using a joint Dice-CE loss. Checkpoints were selected based on mJSW mean absolute error (MAE) to prioritize geometric fidelity. For each compartment $c \in \{\text{med}, \text{lat}\}$, 16 equidistant landmarks along the femoral boundary were sampled, applying endpoint trimming to remove margin artifacts. The compartment minimum width is defined as:
\begin{equation}
\mathrm{mJSW}_{c} = \min_{i \in \{1,\dots,16\}} d\left(f_{i}^{c}, t_{i}^{c}\right),
\label{eq:mjsw}
\end{equation}
where $d(\cdot,\cdot)$ denotes the Euclidean distance between femoral ($f_{i}^{c}$) and tibial ($t_{i}^{c}$) points. The narrowing level is normalized against a healthy reference:
\begin{equation}
\mathrm{JSN\_rate}_{c} = 100 \times \left(1 - \frac{\mathrm{mJSW}_{c}}{\mathrm{mJSW}_{\mathrm{KL0}}}\right),
\label{eq:jsnrate}
\end{equation}
and compartmental structural asymmetry is calculated as follows:
\begin{equation}
\text{Asymmetry} = \frac{\left|\mathrm{mJSW}_{\mathrm{med}} - \mathrm{mJSW}_{\mathrm{lat}}\right|}{\mathrm{mJSW}_{\mathrm{med}} + \mathrm{mJSW}_{\mathrm{lat}}}.
\label{eq:asymmetry}
\end{equation}
The resulting 22-dimensional JSN sub-vector encodes both compartmental mJSW values, a 16-point spatial profile, both JSN rates, a medial-to-lateral JSW ratio, and the asymmetry score.

\subsubsection{Stage 2: Osteophyte ROI Extraction and Ordinal Grading.}
Anatomical patches ($140 \times 140$ pixels) at four sites (medial/lateral femur, medial/lateral tibia) were isolated via a three-tier cascade using Stage~1 landmarks, an anatomical heuristic fallback, or fixed geometric crops. Ordinal grading was performed by a multi-task SE-ResNet-50~\cite{hu2019squeezeandexcitation} with four site-specific multilayer perceptron (MLP) heads trained via ordinal cross-entropy, applying targeted fine-tuning to the lateral femur and medial tibia branches. Grad-CAM heatmaps~\cite{selvaraju2020grad} were generated solely for post-hoc visualization and excluded from downstream processing. The resulting 10-dimensional osteophyte sub-vector encodes the four site-specific grades (0--3), global burden (sum of site grades), maximum site grade, and four directional burden aggregations (medial vs. lateral, femoral vs. tibial).

\subsubsection{Stage 3: Subchondral Sclerosis ROI Extraction and Binary Classification.}
Subchondral ROIs immediately inferior to the tibial plateau were extracted from the Stage~1 mask following Bayramoglu et al.~\cite{bayramoglu2020adaptive}. A hybrid classifier fused handcrafted texture descriptors (multi-scale LBP histograms at radii 1, 2, 3; GLCM statistics; fractal dimensions; per-compartment intensity profiles) with a parallel EfficientNet-B0 image branch via a two-layer MLP. A binary scheme (absent vs. present) was deployed to mitigate convergence instability caused by low mild-sclerosis sample volume, optimizing the threshold at $\tau = 0.4227$ to maximize macro $F_1$. The 18-dimensional sclerosis sub-vector comprises binary compartment grades, localized intensity profiles, fractal dimensions, five GLCM metrics per compartment, and down-sampled LBP entropy features.

\subsubsection{Stage 4: Feature Aggregation and KL Classification.}
The three intermediate structural sub-vectors are concatenated and $z$-score normalized into a unified 50-dimensional feature vector:
\begin{equation}
\mathbf{x} = \bigl[\mathbf{x}_{\mathrm{JSN}}^{(22)} \;\|\; \mathbf{x}_{\mathrm{OSP}}^{(10)} \;\|\; \mathbf{x}_{\mathrm{SCL}}^{(18)}\bigr] \in \mathbb{R}^{50},
\label{eq:featurevec}
\end{equation}
supporting the two path classification framework. \textbf{Path~A (Audit Path)} leverages an XGBoost classifier~\cite{chen2016xgboost} (300 estimators, max depth 6, lr $=0.05$, $L_1$/$L_2$ regularization) trained via 5-fold cross-validation exclusively on $\mathbf{x}$ to extract explanatory SHAP attributions~\cite{NIPS2017_8a20a862}. The deployed model, \textbf{Path~B (Deployment Path)}, implements a ConvNeXt Hybrid architecture, concatenating a 768-dimensional visual encoding from a ConvNeXt-Small backbone~\cite{liu2022convnet} with a 256-dimensional MLP projection of $\mathbf{x}$. This fused vector feeds a downstream MLP for five-class ordinal prediction via soft-label cross-entropy ($\sigma = 0.5$). A web-based Gradio interface exposes these feature-level outputs alongside interactive measurement overlays, processing a single raw radiograph in under 10 seconds.

\section{Experiments and Results}
We report results module by module, beginning with the JSN segmentation and measurement performance and proceeding through osteophyte grading, sclerosis classification, and the final KL grade prediction with its accompanying ablations.

\subsection{JSN Segmentation and Measurement Performance}
\begin{figure*}[!t]
\centering
\includegraphics[width=\textwidth]%
  {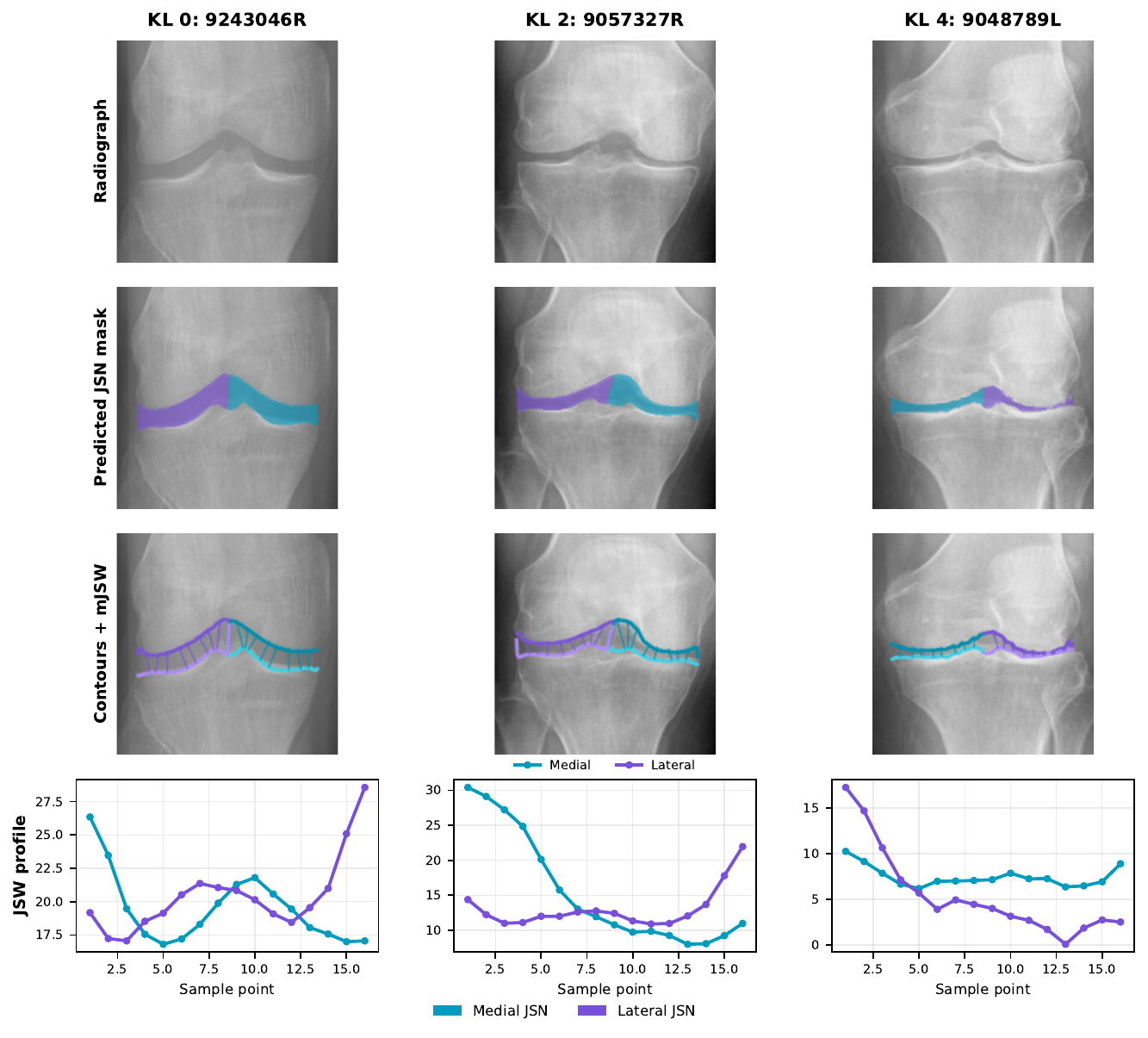}
\caption{JSN module outputs for representative KL grades 0, 2, and 4.}
\label{fig:jsn}
\end{figure*}

Evaluated on the 60-image subset, the expert-supervised U-Net++ checkpoint achieved $\text{Dice} = 0.8904$ and $\text{HD95} = 1.4734\text{ px}$, matching the TransUNet benchmark ($\text{Dice} = 0.889$) of Guo et al.~\cite{guo2025predicting}. An $\text{ICC} = 0.8699$ indicates excellent mJSW reproducibility, exceeding the standard validation threshold ($\text{ICC} \geq 0.75$)~\cite{koo2016guideline}. This supervised checkpoint outperformed its self-trained variant ($\text{MAE} = 1.7020$ vs. $1.7217\text{ px}$; $\text{ICC} = 0.8699$ vs. $0.8645$), yielding a final deployed performance of $\text{Dice} = 0.8909$, $\text{HD95} = 1.4964\text{ px}$, $\text{MAE} = 1.7041\text{ px}$, and $\text{ICC} = 0.8674$. Qualitative visualizations are shown in Fig.~\ref{fig:jsn}.

\subsection{Osteophyte Grading Performance}
Osteophyte grading was heterogeneous across anatomical sites (Table~\ref{tab:osteophyte}). The medial femur achieved the highest agreement ($\kappa = 0.5828$, $\text{AUC} = 0.6872$), followed by the medial tibia ($\kappa = 0.4706$, $\text{AUC} = 0.7105$). Conversely, the lateral femur exhibited a severe validation-to-test drop ($\kappa: 0.6660 \to 0.1048$), driven by limited annotation volume and 2D geometric ambiguity on PA projections. Because pseudo-label expansion amplified teacher bias ($\kappa = 0.3791$ vs. $0.3812$), the expert-supervised framework was retained, treating the lateral femur output as a secondary clinical indicator.

\begin{table*}[!t]
\caption{The per-site osteophyte grading performance (manual-label
framework, held-out test set, $n{=}75$ per site).}
\label{tab:osteophyte}
\centering
\begin{tabular*}{\textwidth}{@{\extracolsep{\fill}}llcccc}
\toprule
Site & Strategy & Val $\kappa$ & Test $\kappa$ & Test Bal.\ Acc. & Test AUC \\
\midrule
Medial femur  & Multitask    & \textbf{0.5707} & \textbf{0.5828} & 0.4427 & 0.6872 \\
Lateral femur & Refined site & 0.6660 & 0.1048 & 0.3228 & 0.5623 \\
Medial tibia  & Refined site & 0.5200 & 0.4706 & 0.4539 & \textbf{0.7105} \\
Lateral tibia & Multitask    & 0.6652 & 0.3665 & \textbf{0.4551} & 0.6773 \\
\midrule
Mean          & ---          & ---    & 0.3812 & 0.4186 & 0.6593 \\
\bottomrule
\end{tabular*}
\end{table*}

\subsection{Subchondral Sclerosis Classification}
\begin{figure*}[!t]
\centering
\includegraphics[width=\textwidth]{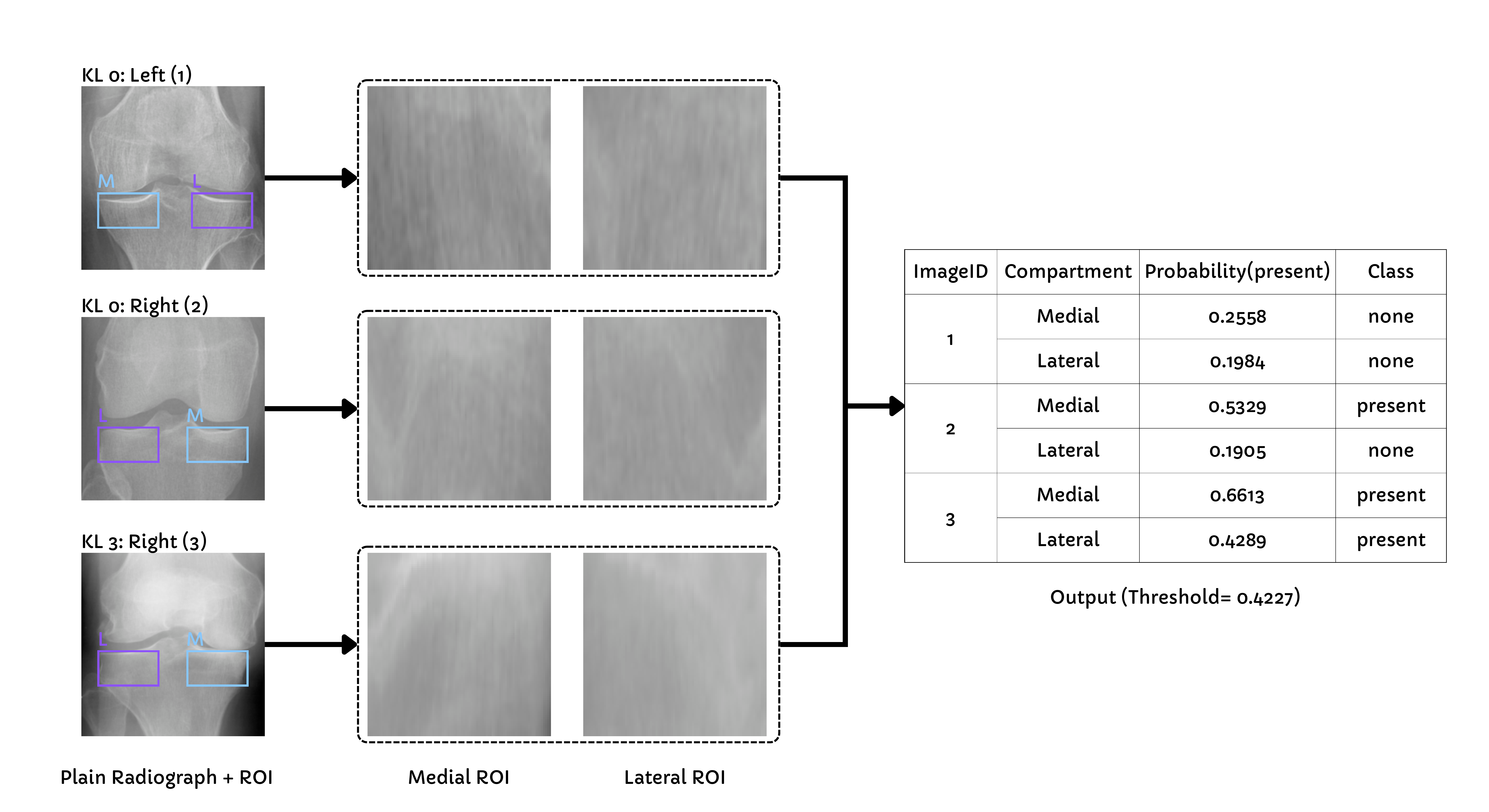}
\caption{Sclerosis module ROI extraction and classification for representative cases.}
\label{fig:sclerosis}
\end{figure*}

On the annotated validation subset ($n = 150$ per split, $\tau = 0.4227$), the hybrid classifier achieved validation macro $F_1 = 0.6212$ ($95\%$~CI: $0.5390$--$0.6866$) and $\text{AUC} = 0.6709$ ($0.5867$--$0.7552$). On the test partition, it yielded macro $F_1 = 0.5785$ ($0.4977$--$0.6582$) and $\text{AUC} = 0.6114$ ($0.5245$--$0.7003$), outperforming the strongest classical baseline (Extra Trees: test macro $F_1 = 0.5696$, $\text{AUC} = 0.6204$). Global accuracy and balanced accuracy tracked within $0.01$ across both splits. Representative ROIs are in Fig.~\ref{fig:sclerosis}. Modest absolute performance (AUC CI lower bound $\approx 0.52$) is analyzed in Section~\ref{sec:limitations}.

\subsection{KL Grade Classification}
Path~B reached test $\text{QWK} = 0.8436$ (validation $\text{QWK} = 0.7966$), accuracy $0.6636$, macro $F_1 = 0.6693$, and $\text{AUC} = 0.9017$ ($95\%$~CI: $0.8920$--$0.9104$). Path~A, restricted to the 50-dimensional feature vector, reached test $\text{QWK} = 0.6294$ (5-fold cross-validation: $0.6467 \pm 0.0109$), accuracy $0.5399$, macro $F_1 = 0.5238$, and $\text{AUC} = 0.8046$ ($0.7922$--$0.8169$). This performance gap reflects Path~A's constrained feature summary rather than an intrinsic interpretability cost. Errors were heavily concentrated at KL~1 (Fig.~\ref{fig:pathb_confusion}, top-left), where recall fell to $42\%$ due to misclassification as KL0 ($35\%$) or KL2 ($21\%$). Recall remained robust across all other grades ($\text{KL0} = 77\%$, $\text{KL2} = 60\%$, $\text{KL3} = 76\%$, $\text{KL4} = 88\%$), with confusions strictly limited to adjacent tiers.

\subsubsection{Path A: Feature-Importance Audit.}
\begin{figure*}[!t]
\centering
\includegraphics[width=0.92\textwidth]{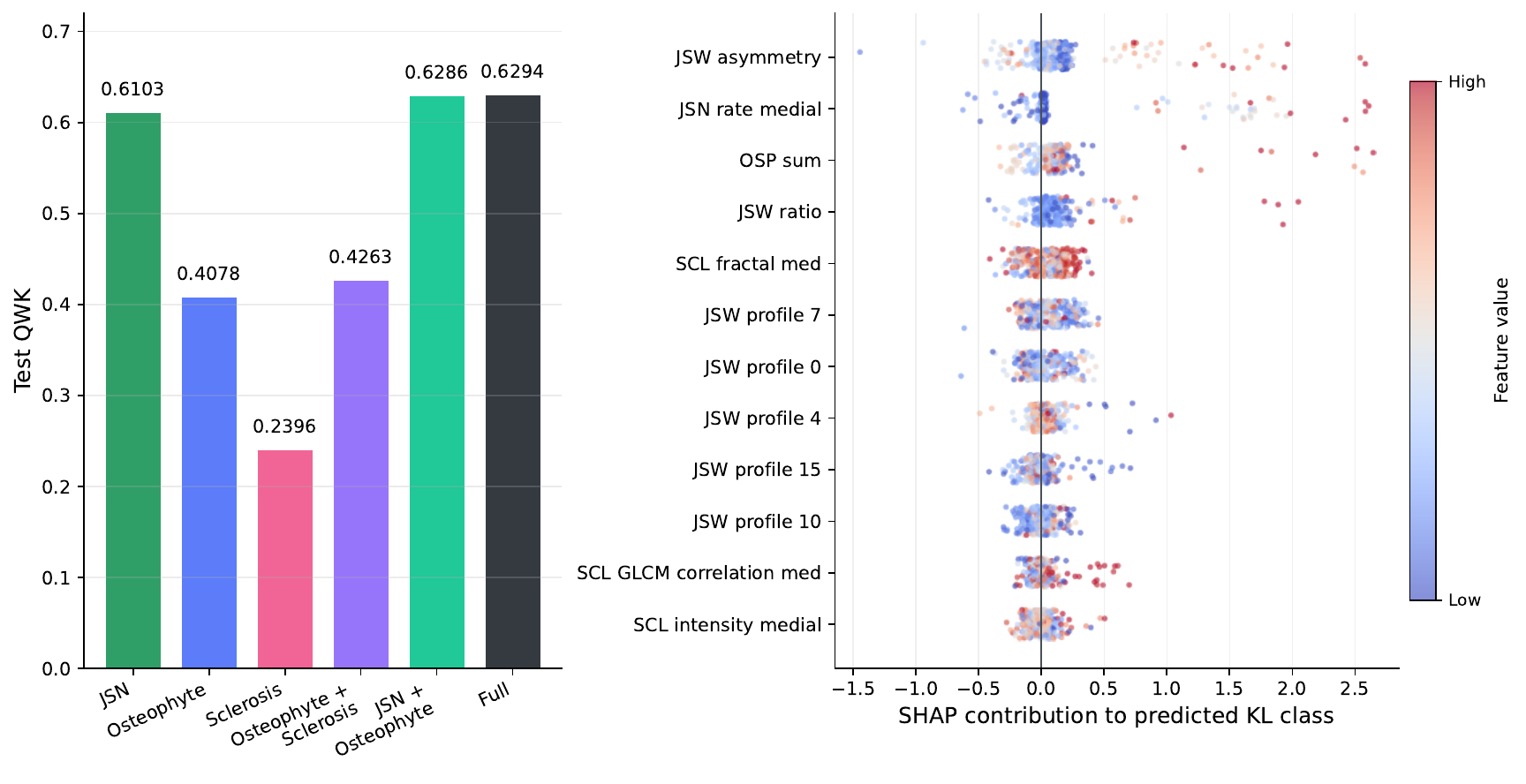}
\caption{Path~A feature-family ablation performance in QWK (left) and global SHAP feature attributions evaluated across the independent test partition (right).}
\label{fig:ablation}
\end{figure*}

Path~A feature importance was heavily dominated by JSN parameters ($\text{QWK} = 0.6103$; Fig.~\ref{fig:ablation}), with osteophytes adding a modest increment ($\Delta\text{QWK} = +0.0183$) and sclerosis contributing marginally ($+0.0008$). Global SHAP attributions independently confirmed this hierarchy, ranking JSW asymmetry and medial JSN rate as the two highest-weight individual features. This clinical weighting (dominant JSN, complementary osteophytes, and marginal sclerosis) precisely mirrors the diagnostic logic of the original KL scale.

\subsubsection{Path B: Inference-Time Reliance of the Deployed Predictor.}
\begin{figure*}[!t]
\centering
\includegraphics[width=0.95\textwidth]{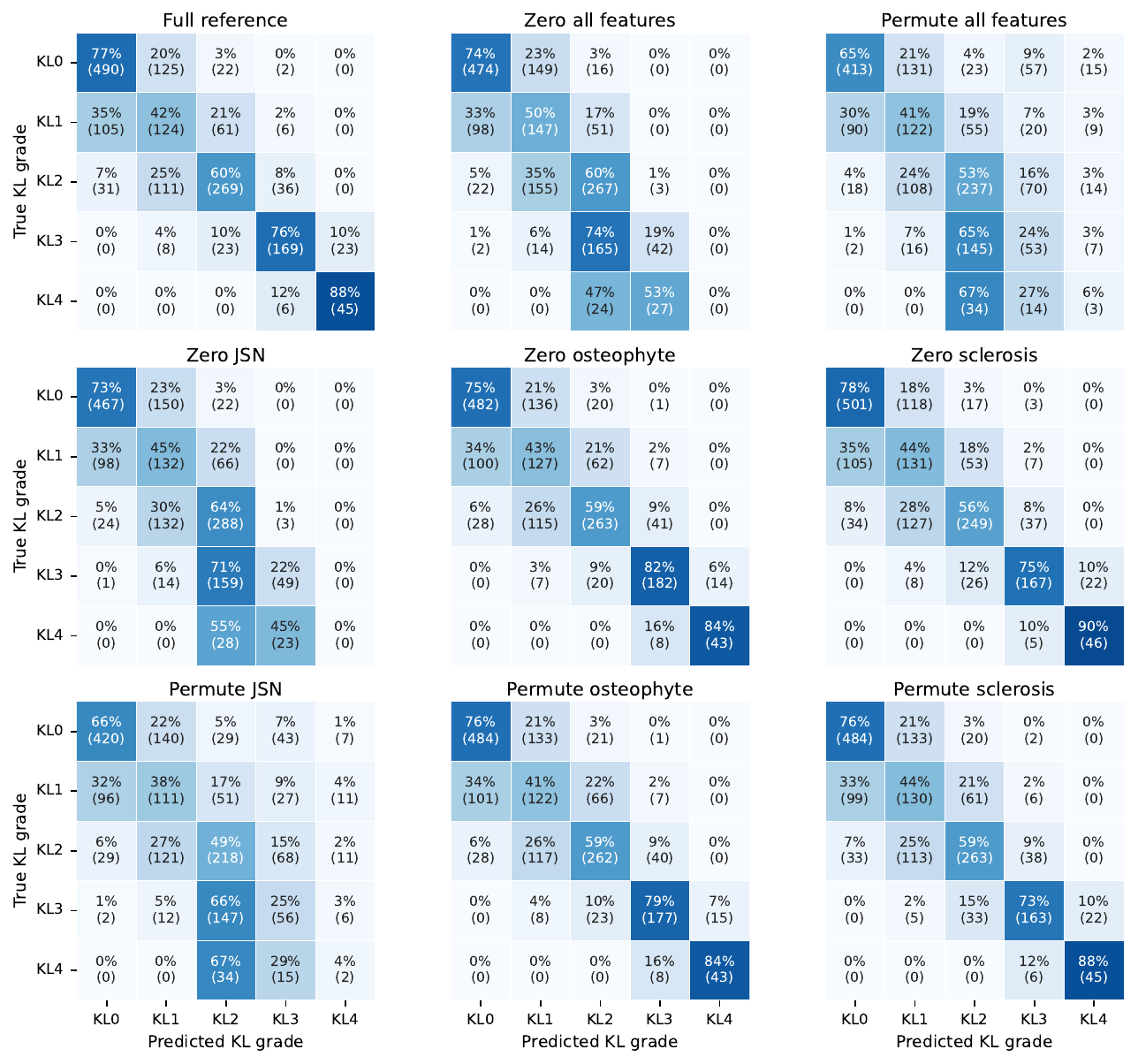}
\caption{Grade-resolved ConvNeXt hybrid confusion matrices under inference-time ablation ($n = 1,656$). Inner cells indicate row-normalized recall values with raw counts in parentheses. Columns show performance changes when JSN, osteophyte, and sclerosis evidence are independently disrupted.}
\label{fig:pathb_confusion}
\end{figure*}

To verify structured feature reliance without network retraining, we applied inference-time zeroing and feature permutation. Zeroing the full feature vector lowered QWK by $0.098$ while permutation dropped it by $0.284$, confirming tight feature-to-image semantic alignment rather than static covariate exploitation (Table~\ref{tab:pathb_ablation}). This effect was driven almost entirely by JSN ($-0.103$ for zeroing, $-0.259$ for permutation), while osteophyte and sclerosis interventions moved QWK by less than $0.01$. Grade-resolved matrices (Fig.~\ref{fig:pathb_confusion}) show that JSN removal collapses advanced recall (KL3 from $76\%$ to $19\%$; KL4 from $88\%$ to $0\%$) while leaving early grades unaffected, demonstrating alignment with clinical criteria where advanced stages are defined by progressive joint space loss. Random permutation additionally injected spurious confidence (misclassifying $9\%$ of KL0 cases as KL3), confirming an authentic joint feature--image representation.

\begin{table*}[!t]
\caption{Inference-time structured-pathway ablation of the
ConvNeXt predictor (Path B) on the held-out test set
($n{=}1{,}656$), with $\Delta$QWK reported relative to the
full-model baseline of 0.8436. Grade-resolved behaviour for
each row is shown in Fig.~\ref{fig:pathb_confusion}.}
\label{tab:pathb_ablation}
\centering
\small
\begin{tabular*}{\textwidth}{@{\extracolsep{\fill}}llcccc}
\toprule
Intervention & Feature group & Dims & Test QWK & $\Delta$QWK & Test AUC \\
\midrule
--- (full)              & none        & 0  & \textbf{0.8436} & \phantom{$-$}0.0000 & 0.9017 \\
\midrule
Zero                    & All         & 50 & 0.7453 & $-0.0983$ & 0.8543 \\
Permute                 & All         & 50 & 0.5593 & $\textbf{-0.2843}$ & 0.7685 \\
\midrule
Zero                    & JSN         & 22 & 0.7410 & $-0.1027$ & 0.8538 \\
Permute                 & JSN         & 22 & 0.5851 & $-0.2586$ & 0.7672 \\
Zero                    & Osteophyte  & 10 & 0.8392 & $-0.0044$ & \textbf{0.9024} \\
Permute                 & Osteophyte  & 10 & 0.8359 & $-0.0077$ & 0.8999 \\
Zero                    & Sclerosis   & 18 & 0.8346 & $-0.0090$ & 0.8995 \\
Permute                 & Sclerosis   & 18 & 0.8343 & $-0.0093$ & 0.8971 \\
\bottomrule
\end{tabular*}
\end{table*}

\section{Discussion}
We interpret the ablation findings in light of the framework's clinical motivation, evaluate the interpretability evidence they provide, and identify the principal limitations of this study.

\subsection{Clinical Interpretation and Feature-Level Evidence}
The ablation confirms a JSN-dominant reliance hierarchy matching clinical KL evaluation. Since grades 2--4 are defined by progressive joint-space loss~\cite{kellgren1957radiological}, JSN is computationally and clinically the most consequential feature. Systematically withdrawing JSN evidence forces the predictor to downgrade severe cases: KL4 recall collapses from $88\%$ to $0\%$ and KL3 from $76\%$ to $19\%$ (Fig.~\ref{fig:pathb_confusion}). This carries severe clinical implications because surgical pathways typically gate total knee arthroplasty at $\text{KL} \geq 3$~\cite{goh2023patients}. Therefore, misgrading a knee as 2 instead of 3 could deny an arthroplasty referral. Aligning the predictor's primary failure mode with this pathognomonic finding confirms that the final grade is anchored firmly in severe disease indicators rather than an incidental correlate.

Per-site osteophyte features add a complementary discriminative signal, whereas subchondral sclerosis functions as a confirmatory late-stage sign. Zeroing the sclerosis sub-vector leaves the grade-resolved matrices essentially unchanged (Fig.~\ref{fig:pathb_confusion}), echoing radiological practice wherein sclerosis contributes minimal diagnostic weight once definitive JSN and osteophyte evidence are established~\cite{karim2021deepkneeexplainer}. The predictor's reliance hierarchy (decisive JSN, complementary osteophytes, and confirmatory sclerosis) thus directly mirrors established KL diagnostic logic. It is this structural correspondence, rather than raw classification accuracy alone, that allows a non-subspecialist clinician to confidently trust or challenge an automated grade against the raw radiograph.

The lateral femur validation-to-test gap reflects limited localized annotations and 2D geometric ambiguity on standard PA projections. Similarly, the sclerosis performance gap ($\text{AUC} = 0.611$ vs. $0.952$ in Kim et al.~\cite{kim2025classification}) stems from distinct design constraints: a smaller annotated dataset ($n = 150$ vs. $4,019$ images per split), a binary instead of three-class scheme, and omitting CLAHE to preserve raw texture fidelity. At its current scale, this module serves as a necessary architectural baseline ensuring comprehensive three-feature reporting aligned with classical diagnostic definitions.

\subsection{Limitations}
\label{sec:limitations}
First, the $224 \times 224$ px PNG format lacks calibrated DICOM pixel spacing, meaning mJSW values are pixel-based and millimeter conversion requires a user-supplied spatial scale factor. Second, feature-level annotation subsets are relatively small (400--500 images per module), which constrains generalization performance most acutely at the lateral femur site, where the sclerosis test AUC confidence interval lower bound approaches chance. Third, inter-annotator reliability was not formally quantified, annotations were expert-reviewed by two practicing orthopaedic specialists but lack a computed kappa agreement score. Finally, external validation has not yet been conducted, and pipeline robustness to institutional, geographic, and acquisition variability, including the resource-constrained Indonesian clinical contexts motivating this work, remains to be established in future deployment trials.

\section{Conclusion}
Knee-xRAI demonstrates that interpretability in AI-assisted KOA grading can be integrated directly into the diagnostic pipeline via explicit, independently quantified structural features that mirror the KL scale, rather than depending on unreliable post-hoc saliency attributions. Decomposing global severity into three measurable radiographic findings provides clinicians with an auditable chain of evidence rather than an opaque black-box prediction. While a single ConvNeXt hybrid architecture executes the primary grading task, a parallel XGBoost/SHAP model and inference-time ablation studies independently audit and verify feature reliance. This audit confirms that model dependence aligns with clinical logic, concentrating heavily on JSN parameters at advanced severity stages. Ultimately, our principal contribution is providing definitive evidence that an architecture grounded in established radiological workflows can match state-of-the-art performance while offering transparent clinical reasoning.


\bibliographystyle{IEEEtran}
\bibliography{references}


\end{document}